\documentclass[acmtog,nonacm,screen]{acmart}

\AtBeginDocument{%
  }

\setcopyright{acmlicensed}
\copyrightyear{2026}
\acmYear{2026}
\acmDOI{XXXXXXX.XXXXXXX}

\acmJournal{TOG}
\acmVolume{1}
\acmNumber{1}
\acmArticle{}
\acmMonth{8}

\usepackage{algorithm}
\usepackage{algpseudocode}
\usepackage{float}
\usepackage{placeins}
\makeatletter
\def\@mkauthors@i{%
  \def\@currentauthors{}%
  \def\@currentaffiliations{}%
  \global\let\and\@typeset@author@line
  \def\@author##1{%
    \ifx\@currentauthors\@empty
      \gdef\@currentauthors{\@authorfont\MakeUppercase{##1}}%
    \else
      \g@addto@macro{\@currentauthors}{\and\MakeUppercase{##1}}%
    \fi
    \gdef\and{}}%
  \def\email##1##2{\g@addto@macro{\@currentaffiliations}{%
    , {\protect\ttfamily ##2}}}%
  \def\affiliation##1##2{%
    \def\@tempa{##2}\ifx\@tempa\@empty\else
      \ifx\@currentaffiliations\@empty
        \gdef\@currentaffiliations{%
          \setkeys{@ACM@affiliation@}{obeypunctuation=false}%
          \setkeys{@ACM@affiliation@}{##1}%
          \@ACM@resetaffil
          \@affiliationfont##2\@ACM@checkaffil}%
      \else
        \g@addto@macro{\@currentaffiliations}{\and
          \setkeys{@ACM@affiliation@}{obeypunctuation=false}%
          \setkeys{@ACM@affiliation@}{##1}\@ACM@resetaffil
          ##2\@ACM@checkaffil}%
      \fi
    \fi
    \global\let\and\@typeset@author@line}%
  \if@ACM@acmcp
    \advance\hsize by -6pc%
  \fi
  \global\setbox\mktitle@bx=\vbox{\noindent\unvbox\mktitle@bx\par\medskip
    \noindent\addresses\@typeset@author@line
    \par\medskip}%
}
\makeatother
\algnewcommand\algorithmicInput{\textbf{Input:}}
\algnewcommand\Input{\item[\algorithmicInput]}
\algnewcommand\algorithmicOutput{\textbf{Output:}}
\algnewcommand\Output{\item[\algorithmicOutput]}

\usepackage{verbatim}
\usepackage{multirow}

\usepackage[export]{adjustbox} 
\usepackage{tabularx}
\usepackage{booktabs}

\newcommand{\eg}{\textit{e.g.}}
\newcommand{\ie}{\textit{i.e.}}

\acmSubmissionID{1048}

\begin{document}

\title{HarmoHOI: Harmonizing Appearance and 3D Motion for Multi-view Hand-Object Interaction Synthesis}

\author{Lingwei Dang}
\affiliation{%
 \institution{South China University of Technology}
 \city{Guangzhou}
 \country{China}
}
\email{levondang@163.com}

\author{Juntong Li}
\affiliation{%
 \institution{South China University of Technology}
 \city{Guangzhou}
 \country{China}
}
\email{qlzjftm@gmail.com}

\author{Zonghan Li}
\affiliation{%
 \institution{South China University of Technology}
 \city{Guangzhou}
 \country{China}
}
\email{leezhh888@gmail.com}

\author{Hongwen Zhang}
\affiliation{%
 \institution{Beijing Normal University}
 \city{Beijing}
 \country{China}
}
\email{zhanghongwen@bnu.edu.cn}

\author{Liang An}
\affiliation{%
 \institution{Tsinghua University}
 \city{Beijing}
 \country{China}
}
\email{anliang@mail.tsinghua.edu.cn}

\author{Wei Min}
\affiliation{%
 \institution{Shadow AI}
 \city{Beijing}
 \country{China}
}
\email{minwei@yingshen-ai.com}

\author{Yebin Liu}
\affiliation{%
 \institution{Tsinghua University}
 \city{Beijing}
 \country{China}
}
\email{liuyebin@tsinghua.edu.cn}

\author{Qingyao Wu$^\dagger$}
\affiliation{%
 \institution{South China University of Technology}
 \city{Guangzhou}
 \country{China}
}
\email{qyw@scut.edu.cn}
\authorsaddresses{}

\begin{abstract}


    Hand-Object Interaction (HOI) synthesis is a cornerstone for animation production and embodied AI. Despite the strong priors of video foundation models, multi-view consistent HOI synthesis remains challenging due to complex hand motions and occlusions. We present HarmoHOI, a unified diffusion framework that jointly and harmoniously generates synchronized multi-view HOI videos and globally aligned 3D point tracks. Our core insight is that robust multi-view consistency fundamentally requires globally aligned 3D geometry and motion. To this end, we propose a Mixture of Multi-view Diffusion Transformer that co-models RGB videos and 3D point tracks. By representing point tracks as pseudo-videos, we align 3D geometric signals with the 2D latent space of foundation models, thereby minimizing the domain gap and easing adaptation of priors. To further ensure geometry consistency, we introduce Global Motion Aligning Diffusion, which refines coarse point tracks into metric-scale, globally aligned 3D trajectories. HarmoHOI enables on-the-fly co-evolution of 2D appearance and 3D motion during denoising. To overcome the scarcity of multi-view HOI data, we employ a hybrid data curriculum learning strategy that successfully transfers generic priors from single-view data to synchronized multi-view generation. Experimental results show that HarmoHOI achieves state-of-the-art performance in visual quality, motion plausibility, and multi-view geometric consistency.
    Code for this paper are at \url{https://droliven.github.io/HarmoHOI_project}.

\end{abstract}

\begin{teaserfigure}
    \captionsetup{type=figure}
    \includegraphics[width=1.\textwidth]{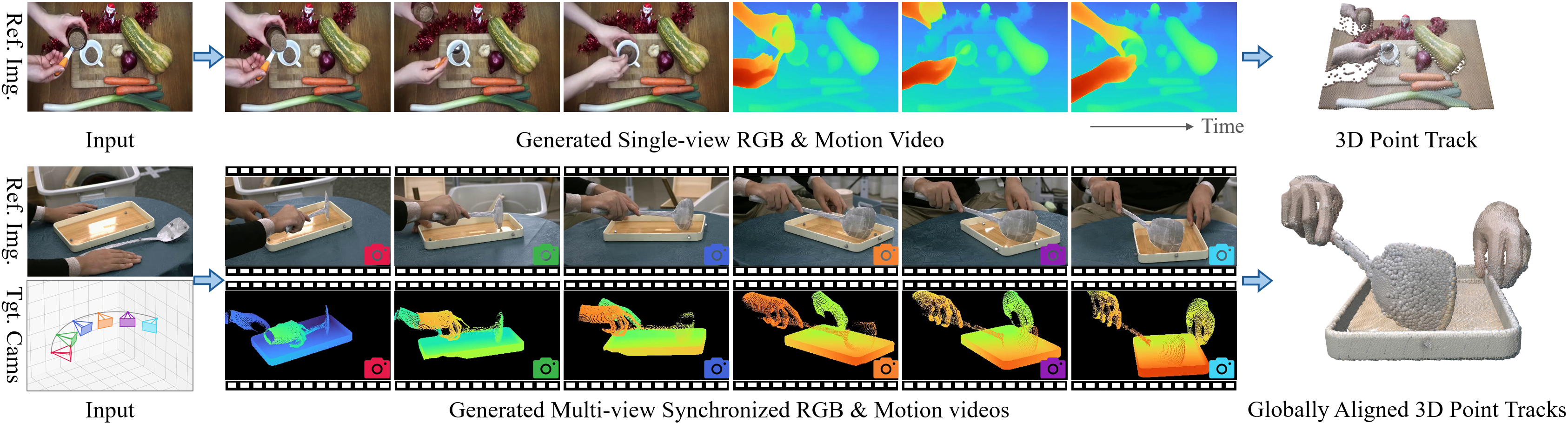}
    \captionof{figure}{
      Our HarmoHOI jointly models the consistency between 2D visual appearance and 3D motion, and learns the synchronization of multi-view epipolar geometry. Consequently, it can not only generate a single-view HOI video and 3D motion from a reference image (top), but also synthesize synchronized multi-view HOI videos and globally aligned 3D point tracks from a reference image and multi-view target camera poses (bottom). The generated outputs exhibit visual realism, motion plausibility, and geometric consistency.
    }
    \label{fig:teaser}
\end{teaserfigure}

\ccsdesc[500]{Computing methodologies~Computer vision}

\keywords{Video Diffusion Model, Hand-Object Interaction, Multi-view Synthesis, Geometry-aware Generation}


\maketitle

\let\thefootnote\relax\footnotetext{
  $\dagger$ Corresponding author
}

\section{Introduction}
\label{sec:intro}

The synthesis of hand-object interaction (HOI) holds significant value for animation production~\cite{xu2024anchorcrafter} and embodied dexterous manipulation~\cite{qin2022dexmv,luo2025being,bharadhwaj2025genact}. Unlike general video generation, HOI involves fine-grained hand movements, frequent extreme self- and mutual occlusions, and complex local deformations~\cite{zhang2025manidext, dang2025svimo, pang2025manivideo, xue2025guiding,li2023object}. These characteristics make the generation of kinematically plausible and multi-view consistent HOIs a highly challenging problem.
Recently, large-scale video foundation models~\cite{brooks2024video,peng2025open,wan2025wan,yang2025cogvideox,kong2024hunyuanvideo,ma2025step} have demonstrated powerful visual priors alongside a certain degree of 3D and physical consistency, offering new technical possibilities for HOI generation.
However, effectively leveraging these video foundation models to generate synchronized multi-view HOI videos while simultaneously obtaining coherent 3D geometry and motion remains an under-explored open problem.

Existing research adapting video models for novel-view or multi-view generation generally falls into three categories.
The first category focuses on camera-controlled novel-view video generation (\eg, Uni3C~\cite{cao2025uni3c}, DaS~\cite{gu2025diffusion}, and MV-Custom~\cite{shin2026mvcustom}). These methods are essentially single-trajectory generative rendering tasks: the models only need to generate a single video along a specified camera path without explicitly verifying whether the same dynamic scene remains geometrically consistent when observed from other viewpoints at the same time. Consequently, they are not naturally designed to guarantee globally consistent multi-view generation in HOI scenarios.
The second category of works (\eg, SV4D 2.0~\cite{Yao_2025_ICCV}, MV-Performer~\cite{zhi2025mv}) attempts to reconstruct multi-view videos from monocular inputs. These are fundamentally generative reconstruction tasks, where temporal dynamics and 3D spatial information are entirely mined from the source video. In contrast, our generation of multi-view HOI processes solely from a single reference image demands significantly higher generative capacity and entails greater uncertainty.
A third line of research directly explores synchronized multi-view video generation (\eg, SynCamMaster~\cite{ICLR2025_9232d474}, CAT4D~\cite{wu2025cat4d}), demonstrating the potential of diffusion models to be adapted into multi-view consistent generators. However, these methods primarily focus on data-driven multi-view visual appearance synchronization. Lacking explicit 3D geometry and motion modeling, they are ill-equipped to handle the fine-grained motions and complex occlusions inherent in HOI scenarios (\textbf{See Supp. Sec.~\ref{tab:add_diss_related} and Tab.~\ref{tab:related_tab}}).

Our core insight is that 2D videos are merely projective snapshots of the 3D physical world: the true key to enabling synchronized multi-view consistency resides in globally aligned 3D geometry and motion awareness. Based on this, we propose \textbf{HarmoHOI}, the first multi-view HOI synthesis framework that harmonizes visual appearance and globally aligned 3D motion within a joint diffusion pipeline. Unlike methods that treat 3D signals as external control conditions~\cite{gu2025diffusion,cao2025uni3c} or rely on post-hoc video reconstruction~\cite{Yao_2025_ICCV,zhi2025mv}, HarmoHOI simultaneously models the consistency between 2D visual appearance and 3D motion, and learns the synchronization of multi-view epipolar geometry during the diffusion generation process. This allows 2D appearance and 3D motion to mutually enhance and co-evolve within a unified generative pipeline.

Specifically, HarmoHOI comprises two key networks.
The first is a Mixture of Multi-view Diffusion Transformer ($M^2$DiT), which builds upon a pre-trained video DiT to construct a dual 2D video and 3D motion joint diffusion model.
By incorporating camera condition embeddings, intra-view spatio-temporal modeling, inter-view geometric attention, and bidirectional mutual modulation, it synchronously generates multi-view consistent HOI videos and 3D motions.
Notably, we choose point tracks as the 3D motion representation. Compared to 2D optical flow~\cite{chefer2025videojam} or 3D keypoints~\cite{dang2025svimo}, point tracks preserve both cross-frame correspondences and 3D geometric information, providing a compact representation that maintains temporal stability and enables robust 3D perception.
To make this representation compatible with video foundation models, we normalize and color-map the depth information into ``motion pseudo videos'', which can be seamlessly encoded by a motion VAE with the same architecture as the vanilla video VAE into a latent space aligned with RGB videos, allowing HarmoHOI to reuse the representational and generative priors of pretrained video models without training a completely separate motion backbone from scratch.

The second key network is the Global Motion Aligning Diffusion (GloMAD). To transform coarse, up-to-scale multi-view 3D point tracks from $M^2$DiT into globally aligned metric-scale 3D motions, we introduce a scale regression mechanism and refine the multi-view trajectories through inter-view geometric attention in a generative diffusion process.
Benefiting from the shared diffusion pipeline between $M^2$DiT and GloMAD, we construct an on-the-fly closed-loop feedback mechanism, enabling the mutual promotion of 2D appearance and 3D motion.

Given the extreme scarcity of synchronized multi-view HOI data, we adopt a hybrid-data progressive curriculum learning strategy. The model is first warmed up on more readily available single-view human-object interaction videos and their pseudo-geometric annotations to learn generic appearance and motion priors. 
Subsequently, synchronized multi-view videos and geometric data are gradually introduced to learn multi-view epipolar geometric consistency. This training strategy not only preserves the visual generalization capabilities that the video foundation model acquired from large-scale data but also ensures the stable injection of multi-view geometric consistency. Experimental results demonstrate that HarmoHOI achieves state-of-the-art performance across multi-view video quality, motion plausibility, and geometric consistency.

In summary, our contributions are threefold:

\begin{itemize}
    \item The first synchronized multi-view joint diffusion framework for HOI video and motion synthesis, achieving high visual quality, motion plausibility, and cross-view consistency.
    \item We integrate a Mixture of Multi-view DiT ($M^2$DiT) for joint appearance-motion modeling with a Global Motion Aligning Diffusion (GloMAD) for 3D trajectory refinement, creating a closed-loop mutual enhancement.
    \item A hybrid data and progressive curriculum training strategy, enabling the model to learn multi-view geometry and motion consistency while maintaining robust visual generalization.
\end{itemize}

\section{Related Work}
\label{sec:related_work}

\textbf{Novel-view and multi-view video synthesis} has three main paradigms: camera-controlled single-trajectory novel-view rendering, multi-view generative reconstruction from monocular inputs, and direct synchronized multi-view video generation. The first paradigm receives the most attention. ReCamMaster~\cite{Bai_2025_ICCV}, Plenoptic VGen~\cite{fu2025plenoptic} and MV-Custom~\cite{shin2026mvcustom} directly generate novel-view videos in a data-driven manner. Other methods~\cite{cao2025uni3c, gu2025diffusion,yang2026neoverse,zhang2026worldstereo,ren2025gen3c, yu2024viewcrafter,jeong2025reangle,liu2025free4d,mark2025trajectorycrafter, bian2025gsdit,shao2024isa4d,Shao2024360degreeHV} adopt a reconstruction-warping-inpainting pipeline with explicit 3D representations. However, limited by single-trajectory generation, they fail to guarantee multi-view consistency.
The second paradigm performs multi-view generative reconstruction from monocular inputs. SV4D~2.0~\cite{Yao_2025_ICCV} is data-driven, while MV-Performer~\cite{zhi2025mv} uses source-view point clouds and normals for geometric guidance. These approaches are ill-posed, as they extract multi-view 3D information from a single video.
The third paradigm~\cite{ICLR2025_9232d474,wu2025cat4d} aims at direct, one-shot synchronized multi-view video generation. Despite efficiency, these methods lack explicit 3D geometric awareness, often producing implausible results. In contrast, our method simultaneously generates multi-view synchronized human-object interactions (HOI) and enhances physical plausibility via joint diffusion of 2D videos and 3D motions, overcoming the key limitations of the aforementioned paradigms.

\textbf{HOI Video models.} Recent video foundation models~\cite{seedance2026seedance,wan2025wan,yang2025cogvideox,kong2024hunyuanvideo} have advanced HOI video generation. Some approaches~\cite{xu2024anchorcrafter,hu2024animate,zhu2024champ} extend UNets with pose guides and appearance networks for pose-controlled synthesis, but their temporal modeling often causes flickering and requires pre-defined pose sequences. More recent works~\cite{chefer2025videojam,dang2025svimo,zhen2025tesseract} leverage Diffusion Transformers (DiT)~\cite{esser2024scaling} for video–motion co-generation to improve physical plausibility. Yet motion representation remains challenging: VideoJam relies on 2D optical flow without explicit 3D awareness, SViMo uses sparse keypoints with limited precision, and TesserAct employs pixel-aligned depth that lacks inter-frame smoothness, while UniMo~\cite{pang2025unimo} jointly models video and 3D motion autoregressively yet remains single-view and human-only. In contrast, our multi-view joint diffusion simultaneously generates 2D videos and metric depth tracks, achieving both 3D awareness and temporal stability.

\textbf{3D HOI reconstruction and generation} primarily relies on high-precision 3D motion capture data~\cite{liu2024taco,chao2021dexycb,zhan2024oakink2,fu2025gigahands,xu2025interact,zhang2022couch,liu2022hoi4d,yang2022oakink,taheri2020grab,fan2023arctic,liu2025hoigen,liu2025core4d}. 
3D HOI reconstruction outputs hand MANO~\cite{MANO_SIGGRAPHASIA_2017} and object meshes, where EasyHOI~\cite{liu2025easyhoi} and Follow My Hold~\cite{aytekin2026follow} take a single image as input, while HOLD~\cite{fan2024hold} and MagicHOI~\cite{wang2025magichoi} take a monocular video as input. In contrast, our method takes a single image as input and outputs overall dense 3D points. As for 3D HOI generation, some works~\cite{cha2024text2hoi,diller2024cg,li2024controllable,zhang2025manidext,lee2024interhandgen,kulkarni2024nifty,liu2024primitive,li2024task,liu2024geneoh} enhance kinematic plausibility by predicting intermediate contact maps or affordances. Others~\cite{xu2024interdreamer,wang2023physhoi,braun2024physically,xu2025intermimic,luo2024omnigrasp} integrate complex physics simulators to improve dynamic realism. However, the limited dataset scale and diversity constrain their generalization. A few approaches~\cite{zhang2025interactanything,zhang2025openhoi} leverage semantic knowledge from multimodal vision-language models (VLMs) to boost HOI generalization, but their multi-stage pipelines are prone to error accumulation.

\section{Method}
\label{sec:method}

\subsection{Preliminary: Basic Video Foundation Model}
\label{sec:preliminary}

Our framework is built upon a pre-trained foundation model for text-to-video generation. It comprises two key components: a spatio-temporal variational autoencoder (VAE)~\cite{kingma2013auto} that compresses the original video $\boldsymbol{V}$ into a more compact latent space $\boldsymbol{z}$, and a Diffusion Transformer (DiT)~\cite{peebles2023scalable} based video generator to synthesize video latents $\hat{\boldsymbol{z}}$. 
Each DiT block incorporates sequential temporal modulation and self-attention among visual tokens, cross-attention between textual and visual tokens, and a feedforward MLP layer.
The model employs the Rectified Flow framework~\cite{esser2024scaling} for noise scheduling and denoising operations. During training, given clean video latents $\boldsymbol{z}_0$, Gaussian noise $\boldsymbol{z}_1 \in \mathcal{N}(\boldsymbol{0}, \boldsymbol{I})$, and a random timestep $t \in [0, 1]$, intermediate noisy latents $\boldsymbol{z}_t$ are constructed through linear interpolation $\boldsymbol{z}_t = (1-t) \cdot \boldsymbol{z}_0 + t \cdot \boldsymbol{z}_1$.
The corresponding ground-truth velocity $\boldsymbol{v}_t$ is defined as: $\boldsymbol{v}_t = \mathrm{d} \boldsymbol{z}_{t} / \mathrm{d} t = \boldsymbol{z}_1 - \boldsymbol{z}_0$.
The model parameterized with $\Theta$ is trained to predict the velocity field using mean squared error loss:
\begin{equation}
    \mathcal{L} = \mathbb{E}_{\boldsymbol{z}_0, \boldsymbol{z}_1, \boldsymbol{c}, t} \left\| \boldsymbol{v}_t - \hat{\boldsymbol{v}}_{\Theta}(\boldsymbol{z}_t, \boldsymbol{c}, t) \right\|_2^2.
\end{equation}
In the inference phase, the framework is processed with iterative denoising:
\begin{equation}
    \boldsymbol{z}_{t-1} = \boldsymbol{z}_t + \Delta t \cdot \hat{\boldsymbol{v}}_{\Theta}(\boldsymbol{z}_t, \boldsymbol{c}, t).
\end{equation}


\subsection{Framework Overview}

Given a single reference image $\boldsymbol{I}_\text{ref} \in \mathbb{R}^{H \times W \times 3}$, multi-view target camera poses $\boldsymbol{\Pi} = \{ \boldsymbol{\pi}_v \}_{v=1}^V$, and a textual prompt $\boldsymbol{P}$, we aim to synthesize synchronized multi-view hand-object interaction (HOI) videos $\boldsymbol{V} \in \mathbb{R}^{V \times T \times H \times W \times 3}$ along with the corresponding motion sequences represented as metric-scale 3D point tracks $\boldsymbol{M} \in \mathbb{R}^{V \times T \times K \times 3}$, where $V$, $T$, $H$, $W$, and $K$ denote the number of viewpoints, temporal frames, height, width, and 3D points, respectively. 

\begin{figure*}[!t]
  \centering
  \includegraphics[width=\linewidth]{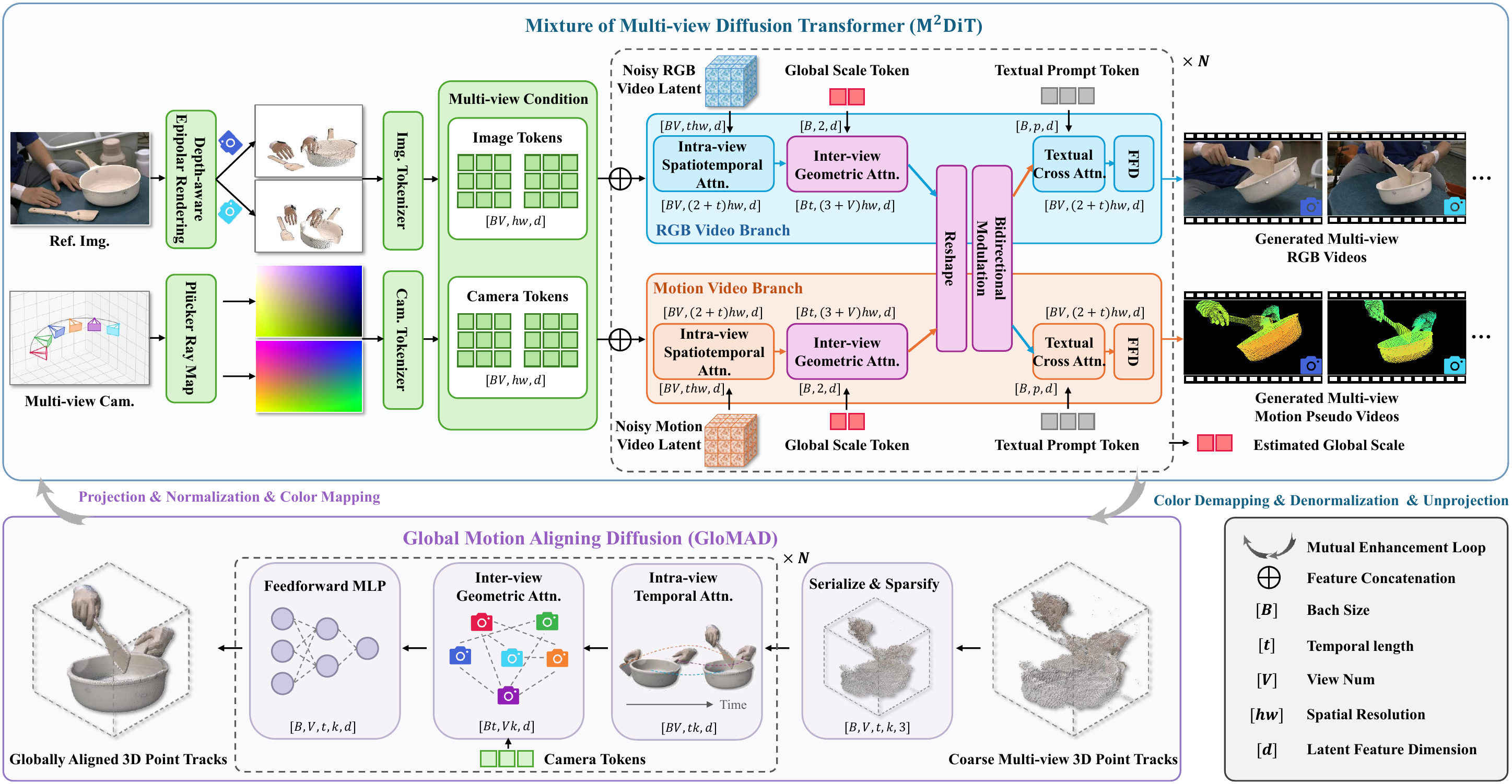}
  \caption{\textbf{Our HarmoHOI framework} comprises two key components: First, the Mixture of Multi-view Diffusion Transformer ($M^2$DiT) generates synchronized multi-view RGB videos, intermediate motion pseudo videos, and global metric scales (Sec.~\ref{sec:m2dit}). Second, the Global Motion Aligning Diffusion (GloMAD) takes the resulting coarse 3D point tracks as a conditioning signal to reconstruct globally aligned point track sequences (Sec.~\ref{sec:glomad}). Furthermore, $M^2$DiT and GloMAD form a closed-loop mutual enhancement cycle during iterative denoising (See Sec.~\ref{sec:cycle}).
  }
  \label{fig:pipeline}
\end{figure*}

To this end, we propose HarmoHOI, an end-to-end multi-view HOI generation framework (Fig.~\ref{fig:pipeline}), whose core consists of two components: (1)~$M^2$DiT jointly generates multi-view RGB and pseudo videos with global metric scales (Sec.~\ref{sec:m2dit}), and (2)~GloMAD refines the resulting coarse point tracks into globally synchronized 3D trajectories (Sec.~\ref{sec:glomad}), forming a mutual enhancement cycle (Sec.~\ref{sec:cycle}). A hybrid-data progressive curriculum strategy (Sec.~\ref{sec:hybrid}) further enables seamless transition from single-view priors to multi-view epipolar consistency modeling.

\subsection{Mixture of Multi-view Diffusion Transformer}
\label{sec:m2dit}
Neither training a 3D point track generative model from scratch nor directly fine-tuning a pretrained video foundation model for motion generation via supervised learning is practical: the former demands massive scaling of 3D motion data to achieve generalization, while the latter suffers from training collapse due to the inherent 2D-3D domain gap. Our strategy is to convert the 3D point track representation into pseudo videos and reuse the video generation model as the backbone for motion generation. This design enables the model to leverage the visual priors of the pretrained video foundation model from the very beginning of training and to progressively adapt to the characteristics of motion generation, thereby yielding more stable training and robust generalization. Most importantly, it allows the 2D and 3D modalities to share a similar latent space, facilitating consistency learning. Consequently, the $M^2$DiT learns to generate multi-view RGB videos $\hat{\boldsymbol{V}}$, pseudo videos of 3D point tracks $\hat{\boldsymbol{M}}_{sv}$, along with the global metric scale $\boldsymbol{s}$. 
A detailed description is provided below.

\textbf{Data Representation and Embedding.} 
Given a source reference image $\boldsymbol{I}_\text{ref}$, optional depth $\boldsymbol{D}_\text{ref}$ and camera parameters $\boldsymbol{\Pi}_\text{ref}$ (estimated via Depth Anything~3~\cite{lin2025depth} when unavailable in training data), and multi-view target camera poses $\boldsymbol{\Pi}$, we apply depth-aware epipolar rendering to warp the reference image to the target viewpoints, yielding multi-view rendered images $\boldsymbol{I} \in \mathbb{R}^{V \times H \times W \times 3}$. 
Concurrently, we unproject the source depth and reproject it into each target camera coordinate system to yield multi-view rendered depth $\boldsymbol{D} \in \mathbb{R}^{V \times H \times W}$, then normalize and colormap the depth into pseudo images $\boldsymbol{I}_\text{pseudo} \in \mathbb{R}^{V \times H \times W \times 3}$. The rendered RGB images $\boldsymbol{I}$ and pseudo images $\boldsymbol{I}_\text{pseudo}$ are then served as the condition for the text-and-image-to-video generation model $M^2$DiT.
The details are summarized in Alg.~\ref{alg:motion_rep}.

\begin{algorithm}[!htbp]
    \caption{Depth-aware epipolar rendering for multi-view condition preparation.}
    \label{alg:motion_rep}
    \begin{algorithmic}[1]
    \Input Reference image $\boldsymbol{I}_\text{ref}$, optional depth $\boldsymbol{D}_\text{ref}$ and camera $\boldsymbol{\Pi}_\text{ref}=\{\boldsymbol{R}_{\text{ref}}; \boldsymbol{T}_{\text{ref}}\}$, target poses $\boldsymbol{\Pi}=\{\boldsymbol{\pi}_v=\{\boldsymbol{R}_v; \boldsymbol{T}_v\}\}_{v=1}^{V}$, intrinsics $\boldsymbol{K}$.
    \Output Multi-view rendered RGB images $\boldsymbol{I}$, pseudo images $\boldsymbol{I}_\text{pseudo}$, and relative Pl\"ucker ray maps $\boldsymbol{\Pi}^\text{rel-plk}$.
    \State $\boldsymbol{P}_\text{ref}^\text{cam} = \text{Unproject}(\boldsymbol{D}_\text{ref}, \boldsymbol{K})$ \Comment{3D points in ref.\ camera}
    \State $\boldsymbol{xy}_{\text{ref}} = \text{Project}(\boldsymbol{P}_\text{ref}^\text{cam}, \boldsymbol{K})$ \Comment{Pixels in $\boldsymbol{I}_\text{ref}$}
    \State $\boldsymbol{P}_\text{ref}^\text{world} = (\boldsymbol{P}_{\text{ref}}^\text{cam} - \boldsymbol{T}_{\text{ref}}^{\text{T}}) \boldsymbol{R}_{\text{ref}}$ \Comment{Lift to world coordinates}
    \For{$v = 1, \cdots, V$}
        \State $\boldsymbol{P}_{v}^\text{cam} = \boldsymbol{P}_\text{ref}^\text{world} \boldsymbol{R}_{v}^{\text{T}} + \boldsymbol{T}_{v}^{\text{T}}$ \Comment{Reproject to target camera}
        \State $\boldsymbol{xy}_{v} = \text{Project}(\boldsymbol{P}_{v}^\text{cam}, \boldsymbol{K})$ \Comment{Project to 2D image plane}
        \State $\boldsymbol{I}^{v}[\boldsymbol{xy}_{v}] = \boldsymbol{I}_\text{ref}[\boldsymbol{xy}_{\text{ref}}]$ \Comment{Render RGB via depth warping}
        \State $\boldsymbol{D}_{v} = \text{Depth}(\boldsymbol{P}_{v}^\text{cam})$ \Comment{Multi-view rendered depth}
        \State $\boldsymbol{D}_{v}^\text{norm} = 1-\text{MinMaxNorm}(\boldsymbol{D}_{v})$ \Comment{Normalize and reverse}
        \State $\boldsymbol{I}_{\text{pseudo}}^{v}[\boldsymbol{xy}_{v}] = \text{Colormap}(\boldsymbol{D}_{v}^\text{norm}) \times 255$ \Comment{Pseudo image}
        \State $\boldsymbol{\pi}_v^\text{rel} = \text{RelativePose}(\boldsymbol{\pi}_v, \boldsymbol{\Pi}_\text{ref})$ \Comment{Convert to relative pose}
        \State $\boldsymbol{\pi}_v^\text{rel-plk} = \text{ToPl\"ucker}(\boldsymbol{\pi}_v^\text{rel})$ \Comment{Compute Pl\"ucker ray map}
    \EndFor
    \State \Return $(\boldsymbol{I}, \boldsymbol{I}_\text{pseudo}, \boldsymbol{\Pi}^\text{rel-plk}) = \{\boldsymbol{I}^{v}, \boldsymbol{I}_{\text{pseudo}}^{v}, \boldsymbol{\pi}_v^\text{rel-plk}\}_{v=1}^{V}$
    \end{algorithmic}
\end{algorithm}

The multi-view rendered RGB images $\boldsymbol{I}$ and pseudo images $\boldsymbol{I}_\text{pseudo}$ are encoded by the video VAE into a latent representation and then tokenized into image tokens $\boldsymbol{f}_{I}, \boldsymbol{f}_{I-\text{pseudo}} \in \mathbb{R}^{V  \times h \times w \times d}$. In parallel, Pl\"ucker ray maps are embedded into camera tokens $\boldsymbol{f}_{\text{cam}} \in \mathbb{R}^{V \times h \times w \times d}$ via a convolutional camera tokenizer. 
Textual prompts $\boldsymbol{P}$ are encoded by the frozen Google umT5 model~\cite{chung2023unimax} and projected to obtain $\boldsymbol{f}_{\text{text}} \in \mathbb{R}^{p \times d}$. 
The RGB video $\boldsymbol{V}$ and the motion pseudo video $\boldsymbol{M}_{sv}$ are encoded into latent codes $\boldsymbol{z}^{V}$ and $\boldsymbol{z}^{M_{sv}}$ respectively.
By applying the forward diffusion process (Sec.~\ref{sec:preliminary}), we obtain noisy latents $\boldsymbol{z}^{V}_t$ and $\boldsymbol{z}^{M_{sv}}_t $, which are subsequently tokenized into video tokens $\boldsymbol{f}_{V}$ and motion tokens $\boldsymbol{f}_{M_{sv}} \in \mathbb{R}^{V \times t \times h \times w \times d}$, where $t$, $h$, and $w$ denote the temporal and spatial resolution, respectively.
Finally, the full hidden states for the video branch $\boldsymbol{f}_{B_V}$ and motion branch $\boldsymbol{f}_{B_M}$ are obtained by concatenating $\boldsymbol{f}_{I}$ and $\boldsymbol{f}_{\text{cam}}$ with $\boldsymbol{f}_{V}$, and  by concatenating $\boldsymbol{f}_{I-\text{pseudo}}$ and $\boldsymbol{f}_{\text{cam}}$ with $\boldsymbol{f}_{M_{sv}}$, respectively, along the temporal dimension.

\textbf{Mixture of Multi-view Diffusion Blocks.} 
We extend the vanilla single-view video DiT blocks into a dual-branch architecture by incorporating inter-view geometric attention modules to capture multi-view correspondence and bidirectional modulation modules for 2D-3D consistency.
Each DiT block consists of sequential intra-view spatiotemporal attention, inter-view geometric attention, cross-branch modulation, text-conditioned cross-attention, and feedforward multi-layer perceptron modules.
Notably, the motion pseudo video undergoes normalization and consequently loses its scale information. To recover this scale, we introduce learnable scale tokens $\boldsymbol{f}_s$ that regress the global metric scale, a quantity crucial for computing globally aligned 3D trajectories.
The operations within each DiT block are formalized as Eq.~\ref{eq:dit_block}:
\begin{equation}
    \label{eq:dit_block}
    \left\{
    \begin{aligned}
        \boldsymbol{f}_{B_{V}}^{\prime} &= \boldsymbol{f}_{B_{V}} + \text{GAttn}_{{V}}\left(\text{STAttn}_{{V}} (\boldsymbol{f}_{I},\boldsymbol{f}_{\text{cam}},\boldsymbol{f}_{V}, \boldsymbol{f}_{s})\right), \\
        \boldsymbol{f}_{B_{M}}^{\prime} &= \boldsymbol{f}_{B_{M}} + \text{GAttn}_{{M}}\left(\text{STAttn}_{{M}} (\boldsymbol{f}_{I-\text{pse.}},\boldsymbol{f}_{\text{cam}},\boldsymbol{f}_{{M_{sv}}}, \boldsymbol{f}_{s})\right), \\
        \boldsymbol{f}_{B_V}^{\prime} &\mathrel{+}= \text{Mod}_{M2V} \left( \boldsymbol{f}_{B_{M}}^{\prime} \right), \quad \boldsymbol{f}_{B_M}^{\prime} \mathrel{+}= \text{Mod}_{V2M} \left( \boldsymbol{f}_{B_{V}}^{\prime} \right), \\
        \boldsymbol{f}_{B_{(\cdot)}}^{\prime \prime} &= \boldsymbol{f}_{B_{(\cdot)}}^{\prime} + \text{FFD}_{{(\cdot)}}\left(\text{CrossAttn}_{{(\cdot)}} (\boldsymbol{f}_{B_{(\cdot)}}^{\prime}, \boldsymbol{f}_{\text{text}})\right),
    \end{aligned}
    \right.
\end{equation}
where ``$\mathrel{+}=$'' denotes the residual connection.
Specifically, for inter-view geometric attention, features are permuted and reshaped into $\left[B V, \left(2+t\right)hw, d\right]$ to enable cross-view token interactions at each timestep, whereas for intra-view spatiotemporal attention, they are reshaped into $\left[B t, \left(3+V\right)hw, d\right]$  to capture intra-view dependencies across frames.

\textbf{Training Objectives.} 
The output video, motion, and scale features from the final DiT block are decoded to yield the multi-view video latent velocity $\hat{\boldsymbol{v}}^{V}$, the motion pseudo video latent velocity $\hat{\boldsymbol{v}}^{M_{sv}}$, and the global metric scale $\hat{\boldsymbol{s}}$. Thus, the $M^2$DiT is optimized with the following objective:
\begin{equation}
    \label{eq:m2dit_loss}
    \mathcal{L}_{\text{$M^2$DiT}} = \mathbb{E} \left[ \| \hat{\boldsymbol{v}}^{V} - \boldsymbol{v}^{V} \|_2^2 + \| \hat{\boldsymbol{v}}^{M_{sv}} - \boldsymbol{v}^{M_{sv}} \|_2^2 + \| \hat{\boldsymbol{s}} - \boldsymbol{s} \|_2^2 \right],
\end{equation}
where $\boldsymbol{v}^{V}$ and $\boldsymbol{v}^{M_{sv}}$ are the ground-truth velocities derived in Sec.~\ref{sec:preliminary}.

\subsection{Global Motion Aligning Diffusion}
\label{sec:glomad}

Using the intermediate motion pseudo video $\hat{\boldsymbol{M}}_{sv}$ and the estimated global depth scale $\hat{\boldsymbol{s}}$, we obtain coarse 3D point tracks $\hat{\boldsymbol{M}}_\text{coarse}$ by reversing Alg.~\ref{alg:motion_rep}, \ie, through color demapping, denormalization, and unprojection.
In practice, however, the resulting 3D trajectories can be inaccurate due to the pixel-wise optimization paradigm.
To address this, GloMAD reframes the task as conditional generation: it takes $\hat{\boldsymbol{M}}_\text{coarse}$ as input and produces globally aligned 3D point sequences $\hat{\boldsymbol{M}}$.
Specifically, GloMAD stacks multiple layers of intra-view temporal attention, camera-conditioned inter-view geometric attention, and feedforward modules, mainly built upon the sparse convolutions of Point Transformer V3~\cite{wu2024point}.
GloMAD is optimized with the following loss function:
\begin{equation}
    \label{eq:glomad_loss}
    \left\{\begin{aligned}
        &\hat{\boldsymbol{v}}^{M} = \mathcal{G}_\text{GloMAD} \left[ {\boldsymbol{M}}_{t}, \hat{\boldsymbol{M}}_\text{coarse}, t \right], \quad \hat{\boldsymbol{M}} = \boldsymbol{M}_{t} - t \cdot \hat{\boldsymbol{v}}^{M}, \\
        &\mathcal{L}_\text{GloMAD} = \mathbb{E}\left[\text{MSE}(\hat{\boldsymbol{v}}^{M}, \boldsymbol{v}^{M}) + \text{D}_{\text{chamfer}}(\hat{\boldsymbol{M}}, \boldsymbol{M})\right].
    \end{aligned}
    \right.
\end{equation}

\subsection{Close-loop Mutual Enhancement Cycle}
\label{sec:cycle}


Inspired by SViMo~\cite{dang2025svimo} and owing to the similar diffusion pipeline between $M^2$DiT and GloMAD, we establish a closed-loop feedback where $M^2$DiT's outputs serve as a coarse motion condition for GloMAD, while GloMAD's globally aligned points are projected and normalized into a motion pseudo video to guide the next-step denoising of $M^2$DiT, as summarized in Alg.~\ref{alg:inf_code}.

\begin{algorithm}[!htbp]
    \caption{Inference process of HarmoHOI.}
    \label{alg:inf_code}
    \begin{algorithmic}[1]
    \Input Text prompt $\boldsymbol{P}$, reference image $\boldsymbol{I}_\text{ref}$, camera poses $\boldsymbol{\Pi}$, VAE $\mathcal{E}$-$\mathcal{D}$, trained $M^2$DiT $\mathcal{G}_{M^2\text{DiT}}$ and GloMAD $\mathcal{G}_\text{GloMAD}$, average depth scale $\bar{\boldsymbol{s}}$.
    \Output Multi-view HOI video $\boldsymbol{V}$ and 3D point tracks $\boldsymbol{M}$.
    \State $(\boldsymbol{I}, \boldsymbol{I}_\text{pseudo}, \boldsymbol{\Pi}^\text{rel-plk}) \leftarrow$ Alg.~\ref{alg:motion_rep}$(\boldsymbol{I}_\text{ref}, \boldsymbol{\Pi})$ \Comment{Multi-view conditions}
    \State $\boldsymbol{c} = \{\mathcal{E}(\boldsymbol{I}), \mathcal{E}(\boldsymbol{I}_\text{pseudo}), \boldsymbol{\Pi}^\text{rel-plk}, \boldsymbol{P}\}$ \Comment{Condition set}
    \State $\boldsymbol{z}^{V}_{T} \!\sim\! \mathcal{N}(\boldsymbol{0}, \boldsymbol{I}),\; \boldsymbol{z}^{M_{sv}}_{T} \!\sim\! \mathcal{N}(\boldsymbol{0}, \boldsymbol{I}),\; \boldsymbol{M}_{T} \!\sim\! \mathcal{N}(\boldsymbol{0}, \boldsymbol{I})$
    \State $\boldsymbol{s} \leftarrow \bar{\boldsymbol{s}}$
    \For{$t = T, \cdots, 1$}
        \State $\tilde{\boldsymbol{M}} = \mathcal{G}_\text{GloMAD}^{\text{no-grad}}\!\big(\boldsymbol{M}_{t},\, \text{Denorm\&Unproj}(\mathcal{D}(\boldsymbol{z}^{M_{sv}}_{t}), \boldsymbol{s}),\, t\big)$
        \State $\tilde{\boldsymbol{z}}^{M_{sv}} = \mathcal{E}\big(\text{Proj\&Norm}(\tilde{\boldsymbol{M}})\big)$ \Comment{Feedback to $M^2$DiT}
        \State $\boldsymbol{z}^{M_{sv}}_{t} \leftarrow \boldsymbol{z}^{M_{sv}}_{t} + \tilde{\boldsymbol{z}}^{M_{sv}}$ \Comment{Inject motion guidance}
        \State $(\hat{\boldsymbol{v}}^{V}, \hat{\boldsymbol{v}}^{M_{sv}}, \hat{\boldsymbol{s}}) = \mathcal{G}_{M^2\text{DiT}}(\boldsymbol{z}^{V}_{t}, \boldsymbol{z}^{M_{sv}}_{t};\, \boldsymbol{c}, t)$
        \State $\boldsymbol{z}^{V}_{t-1} = \boldsymbol{z}^{V}_{t} + \Delta t \cdot \hat{\boldsymbol{v}}^{V}$,\;
              $\boldsymbol{z}^{M_{sv}}_{t-1} = \boldsymbol{z}^{M_{sv}}_{t} + \Delta t \cdot \hat{\boldsymbol{v}}^{M_{sv}}$ \Comment{Sec.~\ref{sec:preliminary}}
        \State $\hat{\boldsymbol{M}}_\text{coarse} = \text{Denorm\&Unproj}\big(\mathcal{D}(\boldsymbol{z}^{M_{sv}}_{t-1}), \hat{\boldsymbol{s}}\big)$
        \State $\hat{\boldsymbol{M}} = \mathcal{G}_\text{GloMAD}\!\big[\boldsymbol{M}_{t},\, \hat{\boldsymbol{M}}_\text{coarse},\, t\big]$ \Comment{Eq.~\ref{eq:glomad_loss}}
        \State $\boldsymbol{M}_{t-1} \leftarrow \hat{\boldsymbol{M}}$,\; $\boldsymbol{s} \leftarrow \hat{\boldsymbol{s}}$
    \EndFor
    \State $\boldsymbol{V} = \mathcal{D}(\boldsymbol{z}^{V}_{0})$,\; $\boldsymbol{M} \leftarrow \boldsymbol{M}_{0}$
    \State \Return $\boldsymbol{V}$, $\boldsymbol{M}$
    \end{algorithmic}
\end{algorithm}

\subsection{Hybrid-data Progressive Curriculum Learning}
\label{sec:hybrid}

\begin{figure}[!htbp]
    \centering
    \includegraphics[width=\linewidth]{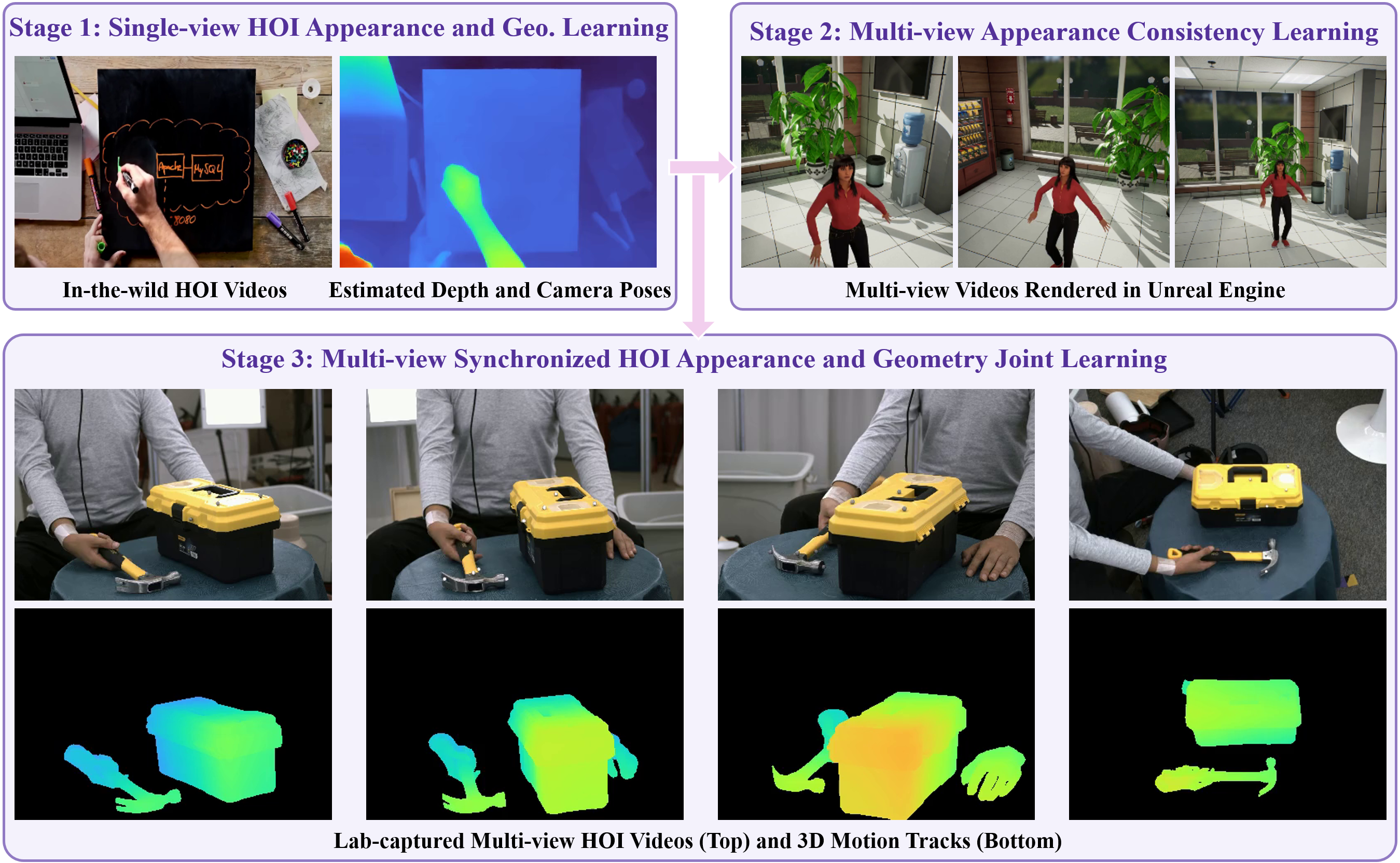}
    \caption{\textbf{Hybrid-data progressive curriculum learning.} HarmoHOI is trained progressively from single-view geometry-aware learning, to multi-view appearance synchronization, and finally to unified multi-view appearance-geometry learning, thereby preserving pretrained visual priors while gradually injecting multi-view geometric consistency.}
    \label{fig:hybrid_data}
\end{figure}

High-quality 3D hand-object interaction (HOI) datasets captured in laboratory settings are increasingly available~\cite{chao2021dexycb,zhan2024oakink2}, yet their scale remains limited, especially for data that simultaneously provides synchronized multi-view videos and accurate 3D motion dynamics. By contrast, large-scale in-the-wild HOI videos~\cite{liu2025hoigen}, though lacking ground-truth annotations, offer rich visual priors and diverse interaction scenarios. To exploit these complementary data sources, we introduce a hybrid-data progressive curriculum learning strategy, as shown in Fig.~\ref{fig:hybrid_data}. 

The training proceeds in three stages with gradually increasing geometric fidelity and multi-view consistency. We first use in-the-wild single-view HOI videos, together with depth maps estimated by a 3D foundation model~\cite{lin2025depth}, to learn the basic correspondence between visual appearance and geometric motion. We then adopt synchronized multi-view videos rendered by Unreal Engine 5 to strengthen cross-view appearance consistency. Finally, we train on high-fidelity laboratory-captured multi-view videos with 3D motion annotations, enabling unified learning of multi-view appearance and geometry. This curriculum preserves the visual priors of pretrained video foundation models while progressively injecting multi-view geometric awareness, leading to more stable training and improved generalization to unseen interaction scenarios.

\section{Experiments}
\label{sec:exp}

\subsection{Settings}

\textbf{Datasets.} Our three-stage hybrid-data pipeline leverages three datasets: HOIGen1M, SynCamVideo, and TACO, as detailed in Tab.~\ref{tab:training_stages}. 
The \textbf{HOIGen1M dataset}~\cite{liu2025hoigen} contains over 1M single-view video clips for HOI video generation, covering diverse interaction types, over 15K objects, over 7K interaction categories, and expressive captions. We estimate depth maps and camera poses for these videos using Depth Anything 3~\cite{lin2025depth}. 
The \textbf{SynCamVideo dataset}~\cite{ICLR2025_9232d474} provides 3.4K distinct dynamic scenes captured by 10 cameras each, yielding 34K videos synthesized from 37 high-quality 3D environment assets, 66 human 3D models, and 93 animations in Unreal Engine 5. We extract video captions using  Qwen 3~\cite{bai2025qwen3}.
Finally, the \textbf{TACO}~\cite{liu2024taco} is among the few public HOI datasets that provide both multi-view videos (12 viewpoints) and high-quality 3D models with accurate poses, comprising over 25K clips. Each interaction follows a tool-action-target triplet structure, covering 20 physical categories (196 instances), 15 action types, and 14 participants. For evaluation, specific objects (\eg, hammer) and actions (\eg, measure) are held out as a generalization test set, with the remaining data split 9:1 into training and validation sets.

\begin{table}[!htbp]
    \centering
    \caption{Hybrid-data Curriculum. Four progressive stages establish image-to-video generation, appearance-geometry correspondence, epipolar consistency, and multi-view HOI synchronization.}
    \label{tab:training_stages}
    \resizebox{\linewidth}{!}{%
        \begin{tabular}{l l l l}
            \toprule
            \textbf{Stage}   & \textbf{Data}  & \textbf{Steps} & \textbf{Architecture} \\
            \midrule
            Warm-up & 10\% HOIGen1M                             & 5K    & Wan backbone \\
            Stage 1 & All HOIGen1M                              & 10K   & + Dual-branch MoT  \\
            Stage 2 & 40\% HOIGen1M + all SynCamVid.           & 10K   & + Inter-view Geo-Attn. \\
            Stage 3 & 10\% HOIGen1M + 20\% SynCamVid. + TACO & 30K   & HarmoHOI \\
            \bottomrule
        \end{tabular}%
    }
\end{table}


\textbf{Metrics.} 
Our method simultaneously generates synchronized multi-view videos and 3D motions. We evaluate video quality based on single-view fidelity and multi-view consistency. For the former, we adopt Subject Consistency and Dynamic Degree from VBench~\cite{huang2024vbench} to assess temporal stability and dynamic quality. Multi-view consistency is measured following SynCamMaster~\cite{ICLR2025_9232d474} using Matching Pixels (via GIM~\cite{shen2024gim}) to quantify pixel alignment and CLIP-Views to evaluate semantic similarity across viewpoints.
For motion evaluation, we consider two lenses: accuracy and plausibility. Accuracy metrics include Chamfer Distance and Motion Smoothness to evaluate geometric fidelity and temporal coherence. Additionally, following GeometryCrafter~\cite{xu2025geometrycrafter}, we use Relative Point Error (RPE) and Percentage of Inliers (PI) to assess point track precision. Multi-view motion consistency is evaluated by applying these same metrics after cross-view reprojection. For motion plausibility evaluation, we compute penetration rate and non-contact rate.



\subsection{Implementation Details.}

Our proposed HarmoHOI extends the pre-trained text-to-video foundation model WAN~2.1-1.3B-T2V~\cite{wan2025wan} to simultaneously generate multi-view synchronized HOI videos and 3D motions from 6 viewpoints. The video resolution is 49 $\times$ 256 $\times$ 384. Our model was trained on 8 NVIDIA A100 GPUs (Refer to \textbf{Supp. Sec.~\ref{sec:add_implement}}). 

\subsection{Comparison with Baselines}

\begin{figure*}[!t]
    \centering
    \includegraphics[width=\linewidth]{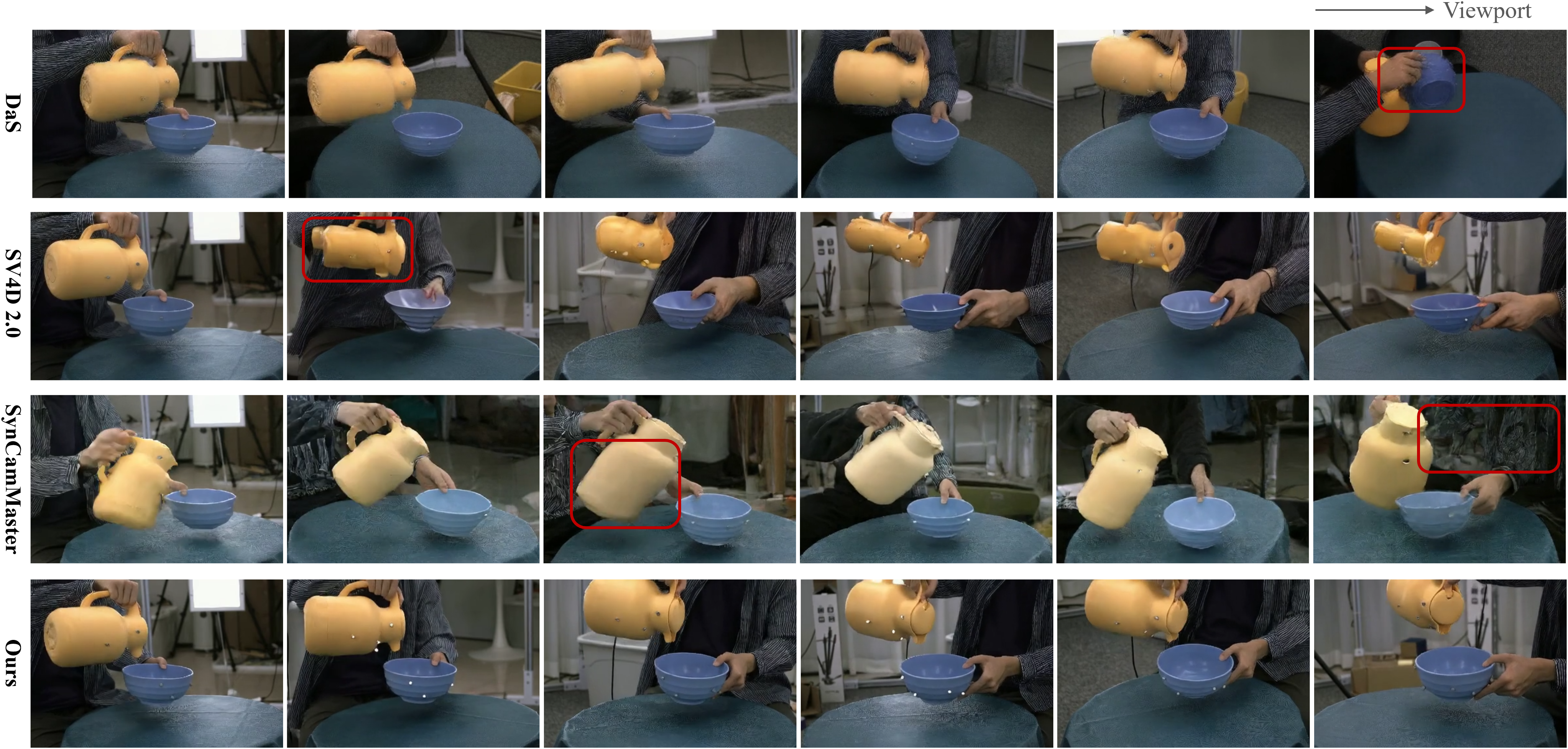}
    \caption{Visualization of the generated multi-view videos from different methods. \textcolor{red}{Red} boxes indicate issues of distortion, object deformation, or multi-view inconsistencies.}
    \label{fig:video_comp}
\end{figure*}

\begin{figure}[!htbp]
    \centering
    \includegraphics[width=\linewidth]{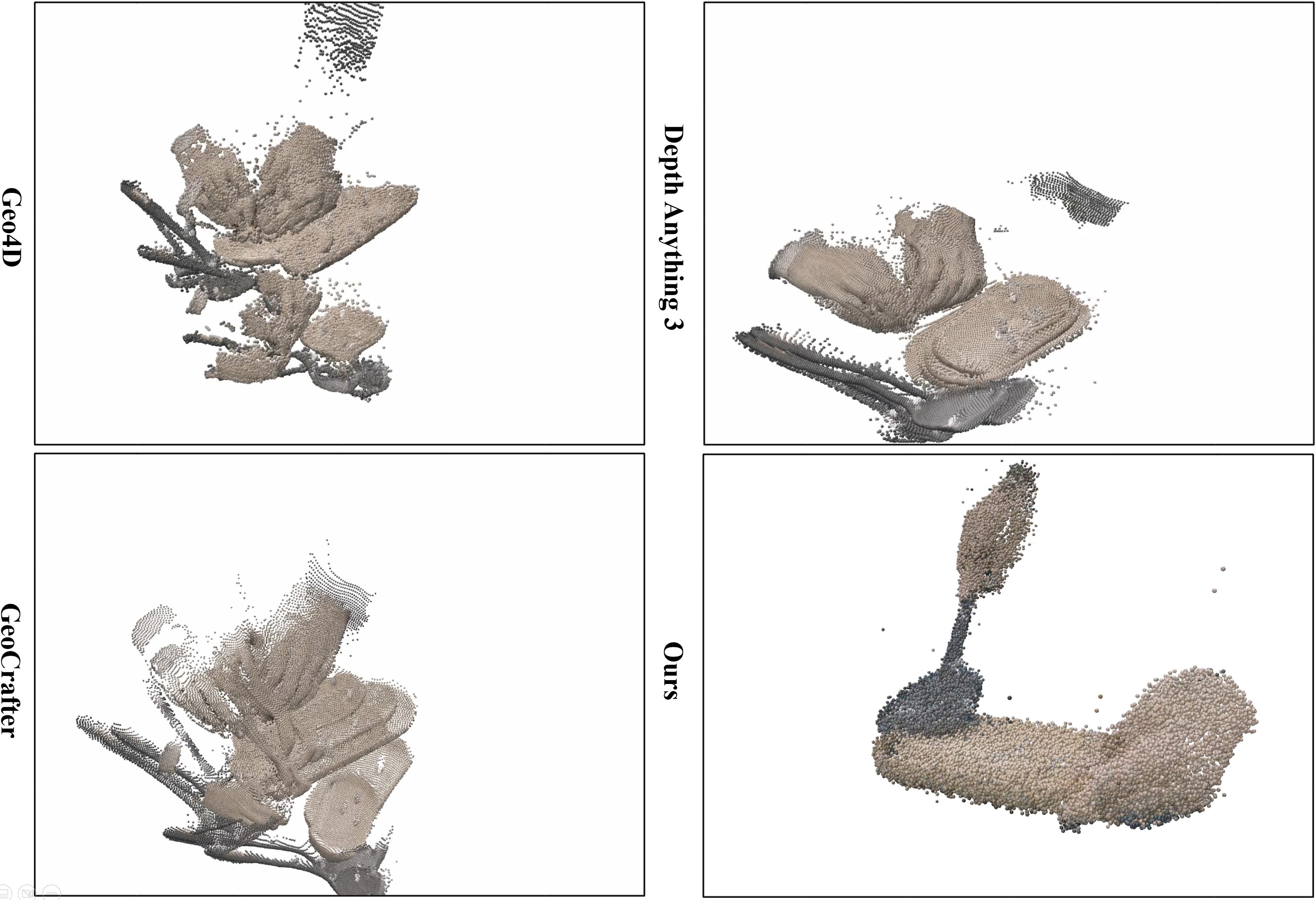}
    \caption{Visualization of multi-view 3D points. Baseline methods exhibit significant misalignment (layer-wise offsets).}
    \label{fig:motions_comp}
\end{figure}

\textbf{Baselines.} For video generation quality evaluation, we compare against single-view video generation models, \ie, WAN~2.1~\cite{wan2025wan} and SViMo~\cite{dang2025svimo}, by generating multi-view videos one by one using reference frames from six different viewpoints. 
We also compare with representative methods from three major categories of novel-view or multi-view video generation, as discussed in Sec.~\ref{sec:intro}, Supp. Sec.~\ref{tab:add_diss_related} and Tab.~\ref{tab:related_tab}. 
The first category is camera-controlled single-trajectory novel-view rendering, such as DaS~\cite{gu2025diffusion}. For this approach, we input a reference video and a target trajectory to generate a video for that specific viewpoint. We repeat this process one by one to generate videos for five viewpoints.
The second category is generative reconstruction from monocular video, exemplified by SV4D~2.0~\cite{Yao_2025_ICCV}. We input a reference video from one viewpoint together with multi-view reference images, and generate videos for five viewpoints simultaneously. 
The third category is synchronized multi-view video generation, represented by SynCamMaster~\cite{ICLR2025_9232d474}. In this case, we input a single reference image and camera parameters for multiple target viewpoints, and generate videos for six viewpoints at the same time. 
For a fair comparison, we fine-tuned all the above baselines on our data using their open-source code.

For 3D motion evaluation, since most existing methods~\cite{jiang2025geo4d,xu2025geometrycrafter,lin2025depth,yuan2025self} are designed for video input to reconstruct the corresponding 3D geometry, and almost no existing method directly generates 3D sequences from a reference image, we compare our image-to-3D generation model with representative single-view video-to-3D reconstruction approaches, \ie, Geo4D~\cite{jiang2025geo4d}, GeometryCrafter~\cite{xu2025geometrycrafter} and Depth Anything 3~\cite{lin2025depth}. 
It should be noted that these methods estimate 3D points one viewpoint at a time, resulting in misalignment issues (layer-wise offsets). Therefore, we post-processed and aligned the multi-view 3D points using the Iterative Closest Point (ICP) algorithm. In contrast, our method directly yields globally aligned 3D point tracks and requires no such transformation.



\textbf{Qualitative and Quantitative Comparison.} 
The comparison results on video generation quality are presented in Tab.~\ref{tab:video_quality} and Fig.~\ref{fig:video_comp}. Our method achieves the best multi-view consistency. However, a single metric provides only a partial view, so multiple metrics and visual demonstrations should be considered together. In particular, WAN~2.1~\cite{wan2025wan} obtains the lowest subject consistency score because its generated videos exhibit severe distortion and poor plausibility. For DaS~\cite{gu2025diffusion}, a novel view generation method, it achieves the second highest object consistency score. Yet the visual demo shows that this is actually due to its weak camera control, which makes the generated novel views similar to the source view video in most cases. Among multi-view generation methods, SV4D~2.0~\cite{Yao_2025_ICCV} achieves the highest dynamic degree score, partly because it uses both the source view video and multi-view reference frames as input. SynCamMaster~\cite{ICLR2025_9232d474} employs implicit camera control without explicit 3D or multi-view visual signals, resulting in relatively weak multi-view consistency.


\begin{table}[!htbp]
    \centering
    \caption{Comparison of video quality. The best and second best results are highlighted with \textbf{bold} and \underline{underlined} fonts. Note that ``SV'', ``NV'' and ``MV'' means single view, novel view and multi-view, respectively.}
    \label{tab:video_quality}
    \resizebox{0.45\textwidth}{!}{\begin{tabular}{c|c|cc|cc}
        \toprule
        \multirow{2.5}{*}{\textbf{Method}} & \multirow{2.5}{*}{\textbf{View}} & \multicolumn{2}{c|}{\textbf{Single-view}} & \multicolumn{2}{c}{\textbf{Multi-view}}  \\ \cmidrule(lr){3-4} \cmidrule(lr){5-6}
            &      & \textbf{Subj. Cons.} & \textbf{Dyn. Deg.} & \textbf{Mat. Pix.} & \textbf{CLIP-V} \\ \midrule
        WAN~2.1  & \multirow{2}{*}{SV}         &  0.5408 &  0.9785        & 246.3         & 71.84  \\
        SViMo     &   &    0.8872 & 0.9634 & 383.2 &   82.45   \\ \midrule
        DaS       & NV       &      \underline{0.9151}  & 0.9713             & \underline{484.7}   & \underline{83.01}    \\ \midrule
        SV4D 2.0    & \multirow{3}{*}{MV}       &  0.8024  & \textbf{0.9931} & 446.8 & 81.33   \\
        SynCamMaster    &        &  0.8657  & 0.8592 & 410.4 & 78.97   \\
        Ours &      &   \textbf{0.9342} & \underline{0.9846}  & \textbf{535.8} & \textbf{83.18}   \\ \bottomrule
        \end{tabular}}
    \end{table}

For 3D motion generation, as shown in Table~\ref{tab:motion_quality} and Fig.~\ref{fig:motions_comp}, our method demonstrates superior performance both quantitatively and qualitatively, whereas the baselines exhibit obvious multi-view inconsistencies (layer-wise offset issue). These gains stem from our method's joint generation of multi-view consistent points, in contrast to the baseline methods that suffer from inherent inconsistencies due to their per-view video reconstruction.





\begin{table}[!htbp]
    \centering
    \caption{Quantitative Comparison of 3D Motions. Best in \textbf{Bold}.}
    \label{tab:motion_quality}
    \resizebox{\linewidth}{!}{\begin{tabular}{c|cccc|cc|cc}
    \toprule
    \multirow{2.5}{*}{\textbf{Method}} & 
    \multicolumn{4}{c|}{\textbf{Accuracy (SV)}} & \multicolumn{2}{c|}{\textbf{Accuracy (MV)}} & \multicolumn{2}{c}{\textbf{Plaussibility}} \\
    \cmidrule(lr){2-5} \cmidrule(lr){6-7} \cmidrule(lr){8-9}
        & \textbf{Chamf.} & \textbf{Smoo.} & \textbf{RPE} & \textbf{PI}  & \textbf{RPE}  & \textbf{PI}  & \textbf{Penetr.} & \textbf{Non-cont.} \\
    \midrule
    Geo4D & 0.0352 & 0.8523 & 21.9 & 64.8 & 79.7 & 2.89 & 0.1275 & 0.19 \\
    GeoCrafter & 0.0315 & 0.8778 & 18.4 & 78.5 & 68.2 & 5.64 & 0.1064 & 0.15 \\
    DA~3 & 0.0241 & 0.9251 & 16.7 & 87.4 & 61.9 & 11.05 & 0.0847 & 0.16 \\
    Ours & \textbf{0.0097} & \textbf{0.9523} & \textbf{14.8} & \textbf{98.6} & \textbf{34.7} & \textbf{34.24} & \textbf{0.0571} & \textbf{0.12} \\
    \bottomrule
    \end{tabular}}
    \end{table}

\textbf{Generalization.}
As described in our dataset protocol, specific objects and actions are held out from TACO for task-level generalization evaluation. Qualitative results on these unseen TACO interactions and in-the-wild inputs are shown in Figs.~\ref{fig:figonlypage_p2}, \ref{fig:figonlypage_p1} and~\ref{fig:figonlypage_p3}, with further examples in the supplementary video. To assess cross-dataset robustness, we further evaluate on a mixed unseen-OOD set of 90 cases sampled from TACO~\cite{liu2024taco}, DexYCB~\cite{chao2021dexycb}, and OakInk2~\cite{zhan2024oakink2}. On this set, HarmoHOI achieves multi-view Matching Pixels of 527.3 and multi-view RPE of 37.1, remaining close to the TACO test-set scores of 535.8 and 34.7 (Tabs.~\ref{tab:video_quality} and~\ref{tab:motion_quality}), indicating robust out-of-distribution generalization.

\begin{figure}[!htbp]
  \centering
  \includegraphics[width=0.95\linewidth]{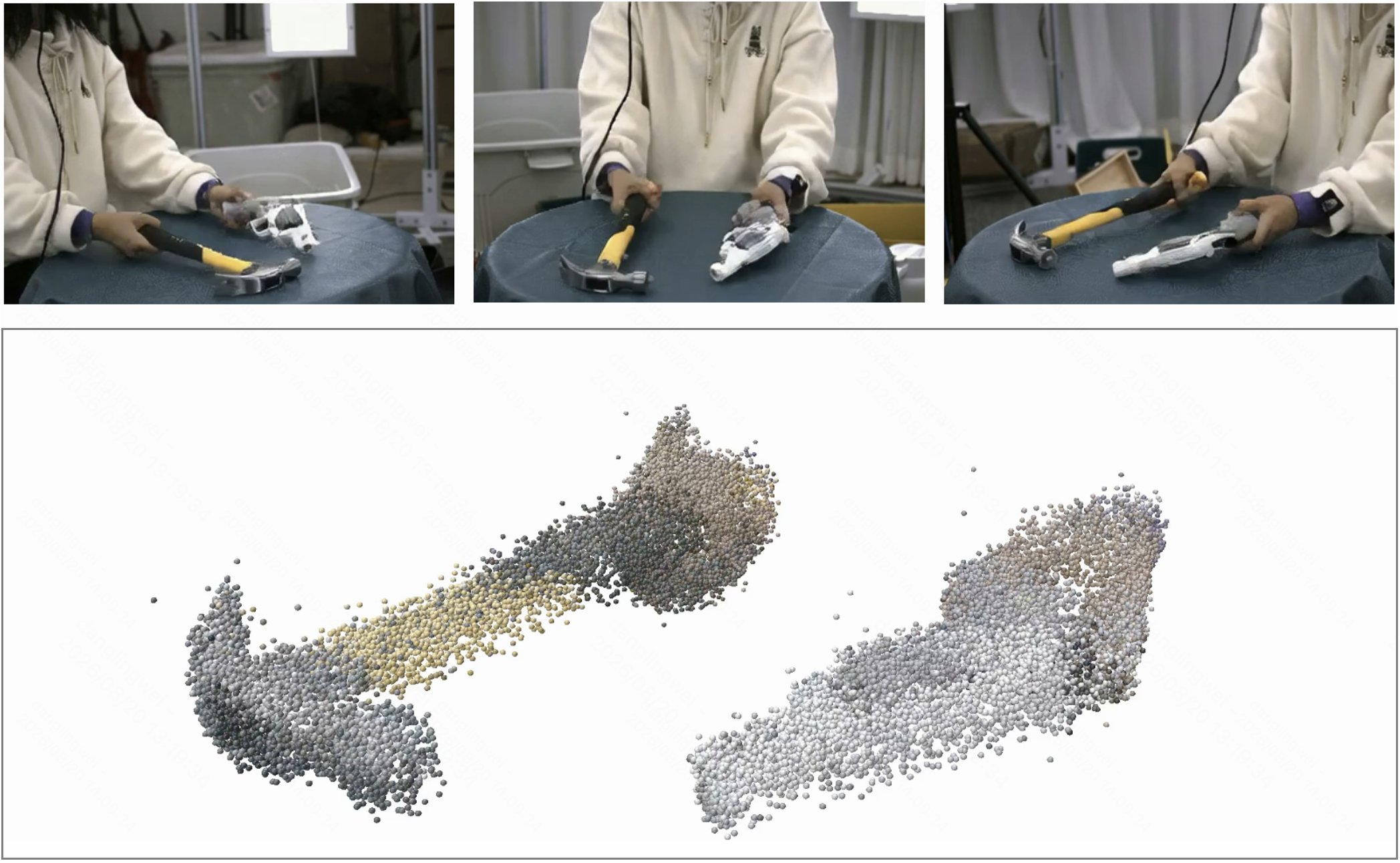}
  \caption{
      Results demonstrating generalization to the UNSEEN HOI task ``use hammer to hit the gun''.
 }
  \label{fig:figonlypage_p2}
\end{figure}

\begin{figure}[!htbp]
  \centering
  \includegraphics[width=\linewidth]{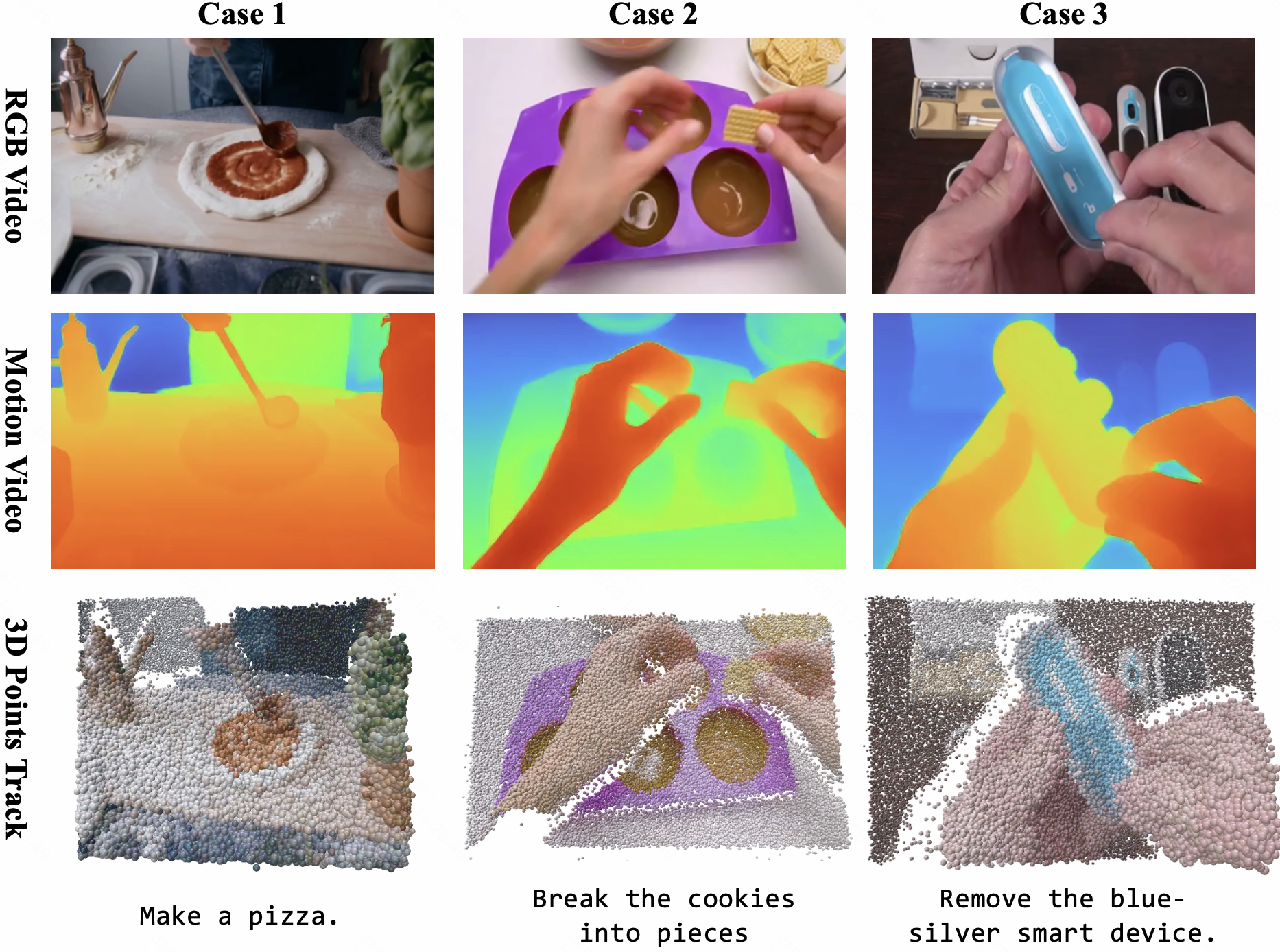}
  \caption{
      In the wild single-view generalization demonstration. From top to bottom are the generated 2D videos, motion pseudo-video, and 3D point tracks. The text prompts are at the bottom.
  }
  \label{fig:figonlypage_p1}
\end{figure}

\begin{figure}[!htbp]
  \centering
  \includegraphics[width=\linewidth]{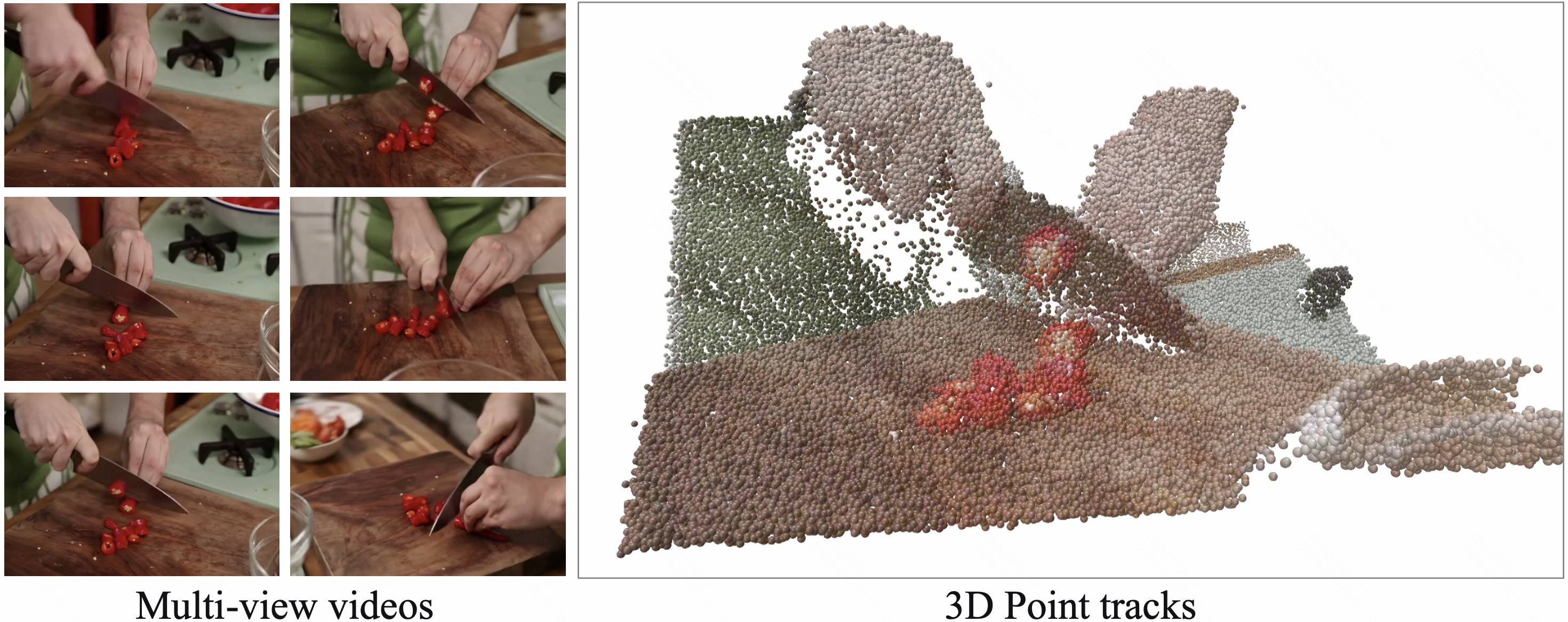}
  \caption{
      Results of in-the-wild multi-view generalization for the HOI task ``slicing red chilies''.
  }
  \label{fig:figonlypage_p3}
\end{figure}

\textbf{Failure Modes.}
While HarmoHOI generalizes well in many settings, residual failures still reflect the coverage of paired multi-view training data. Single-view generation handles diverse in-the-wild cases well, benefiting from HOIGen1M and Depth Anything~3~\cite{lin2025depth} (Fig.~\ref{fig:figonlypage_p1} and the supplementary video). Multi-view generation can degrade under large distribution shifts, since only stage~3 TACO provides paired videos and ground-truth 3D motion. A typical artifact is the layer-wise offset among cross-view 3D point tracks, caused by errors in multi-view metric-scale estimation---also seen in advanced feed-forward reconstructors such as Depth Anything~3.

\begin{table}[!htbp]
    \centering
    \caption{Ablation study on key components of HarmoHOI, including simultaneous multi-view generation (MV), video-motion joint diffusion ($M^2$DiT), motion aligning diffusion (GloMAD), and the hybrid-data training.}
    \label{tab:ablation}
    \resizebox{\linewidth}{!}{\begin{tabular}{cccc|cc}
        \toprule
        \textbf{MV} & \textbf{$M^2$DiT} & \textbf{GloMAD} & \textbf{Hybrid-Data} & \textbf{Mat. Pix. (MV)} & \textbf{RPE (MV)}  \\ \midrule
                           & $\checkmark$            & $\checkmark$            & $\checkmark$              & 335.1              & 73.8          \\
        $\checkmark$                  &              &              &       $\checkmark$    & 438.4              & -             \\
        $\checkmark$                  & $\checkmark$            &              &      $\checkmark$     & 503.7              & 47.6          \\
        $\checkmark$                  & $\checkmark$            & $\checkmark$            &                & 522.5               & 38.4        \\ \midrule
        \multicolumn{4}{c|}{Ours}                                         & \textbf{535.8}     & \textbf{34.6} \\ \bottomrule
        \end{tabular}}
    \end{table}

\subsection{Ablation Study}

We conduct ablation studies to evaluate each component. 
Adapting HarmoHOI to a single-view generation framework severely degrades multi-view consistency and 3D motion quality (Tab.~\ref{tab:ablation}, row 1), as sequential viewpoint generation fails to maintain consistent HOI patterns. 
Removing the video-motion joint diffusion leaves a pure multi-view video model, which improves consistency over the single-view setting but still underperforms HarmoHOI (row 2), resulting in blurriness and deformation. Excluding the GloMAD module and relying only on $M^2$DiT to obtain 3D point tracks via pseudo video denormalization and unprojection substantially reduces motion quality (row 3), since the video foundation model lacks 3D motion training and the VAE introduces information loss. 
Finally, training solely on the TACO dataset without the three-stage hybrid strategy slightly degrades both video and motion consistency (penultimate row).

\section{Conclusion}
\label{sec:conclusion}


In this work, we present a method for synchronized multi-view video and motion co-generation in hand-object interactions. By combining a joint appearance-motion diffusion model with global motion aligning diffusion, we ensure visual realism, plausible motion, and cross-view geometric consistency. A multi-stage hybrid-data curriculum learning strategy further enhances generalization. With only a reference image and a text instruction, the approach is highly accessible and particularly effective under occlusions typical of hand-object interaction.

\textbf{Limitation.}
Paired multi-view HOI video and 3D motion data remain scarce. Even TACO covers only 12 viewpoints, limiting both distribution coverage and view density. A promising next step is to collect or synthesize denser multi-view HOI data and train a 4D Gaussian representation for arbitrary-viewpoint rendering. We believe our framework offers valuable insights for building physics-aware video world models.









\begin{acks}
  This work was supported by National Natural Science Foundation of China (NSFC) 62272172 and GuangDong Basic and Applied Basic Research Foundation 2025B1515120037.
\end{acks}

\bibliographystyle{ACM-Reference-Format}
\bibliography{my_ref}


\clearpage

\appendix
\renewcommand{\shorttitle}{HarmoHOI: Supplementary Materials}

\twocolumn[{%
  \Large\bfseries\centering
  HarmoHOI: Harmonizing Appearance and 3D Motion for Multi-view Hand-Object Interaction Synthesis \\
  ---Supplementary Materials---
  \par\vspace{1cm}
}]
\setcounter{page}{1}
\setcounter{section}{0}

\section{Discussion on Related Works}
\label{tab:add_diss_related}

\begin{figure}[!htbp]
    \centering
    \includegraphics[width=0.8\linewidth]{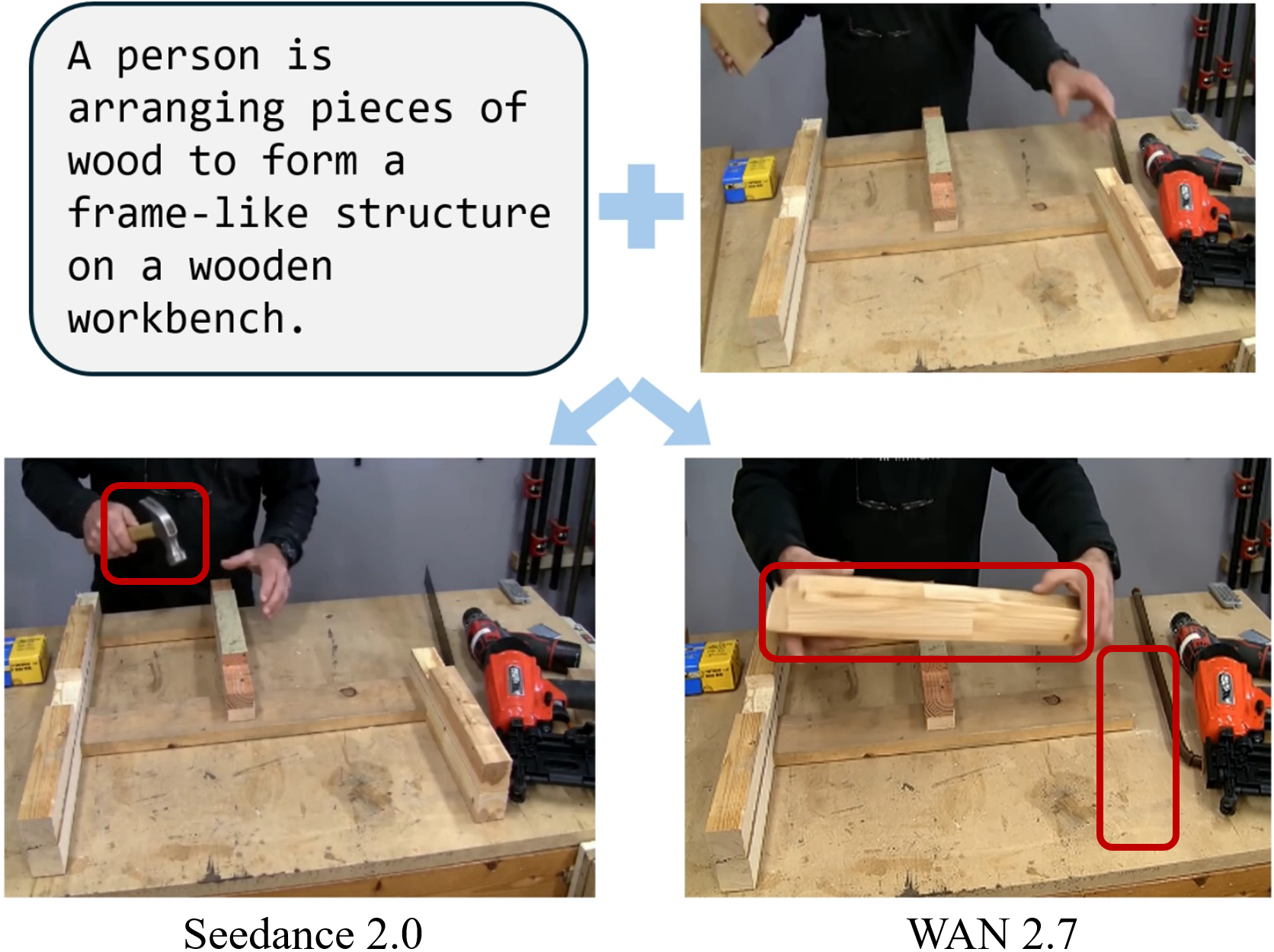}  
    \caption{HOI generation results of top-ranked video models. \textcolor{red}{Red} boxes indicate hallucinations or distortion issues. See the video demonstration in the Appendix for a more vivid demo.}
  \label{fig:seedance}
  \end{figure}

Although video generation models have achieved increasing success in visual quality and dynamic plausibility, challenges remain in generating fine-grained hand object interactions (HOI). 
We tested state-of-the-art video models, including Seedance~2.0~\cite{seedance2026seedance} and WAN~2.7~\cite{wan2025wan}, as shown in Figure~\ref{fig:seedance}. Seedance~2.0 hallucinates a hammer that should not exist, indicating its weak instruction following ability. WAN~2.7 produces unreasonable distortion of the wooden material, showing weak object shape consistency. 
These issues suggest that existing video generation models still have limitations when handling detail rich HOI scenarios. 

Regarding multi-view generation, we have briefly discussed different types of related work on novel view and multi-view generation in Sec.~\ref{sec:intro} and Sec.~\ref{sec:related_work}. We present qualitative and quantitative comparisons of the generation results from representative methods of each category in Sec.~\ref{sec:exp}. To offer a more intuitive visual comparison of these approaches, we show their schematic sketches in Table~\ref{tab:related_tab} and provide a summary comparison of their properties in terms of generative capacity, 3D modeling, and view consistency. This comparison shows that our HarmoHOI, which takes only a single reference image and multi-view target cameras as input, is capable of generating synchronized multi-view hand object interaction videos and 3D motion sequences. It therefore constitutes a framework that simultaneously satisfies high visual realism, 3D modeling capability, and multi-view consistency.


\renewcommand{\tabularxcolumn}[1]{m{#1}}
\newcolumntype{Y}{>{\centering\arraybackslash}X}

\begin{table*}[!t]
    \centering
    \caption{\textbf{Comparison of novel-view / multi-view video synthesis methods.} In the schematics, dark orange and light orange denote mandatory and optional inputs, respectively; green represents the generated visual content (video); and blue indicates the simultaneous generation of both visual content and 3D motion.}
    \label{tab:related_tab}
    \renewcommand{\arraystretch}{1.5} 
    \small 
    
    \begin{tabularx}{\textwidth}{@{} >{\bfseries}l YYYY @{}}
    \toprule
    Category & \textbf{Camera-controlled Novel-view Rendering} & \textbf{Generative Recon. from Monocular Video} & \textbf{Sync. Multi-view Video Generation} & \textbf{Sync. multi-view video and 3D motion generation} \\ \midrule
    
    Schematic  & 
    \includegraphics[width=2.7cm, valign=m]{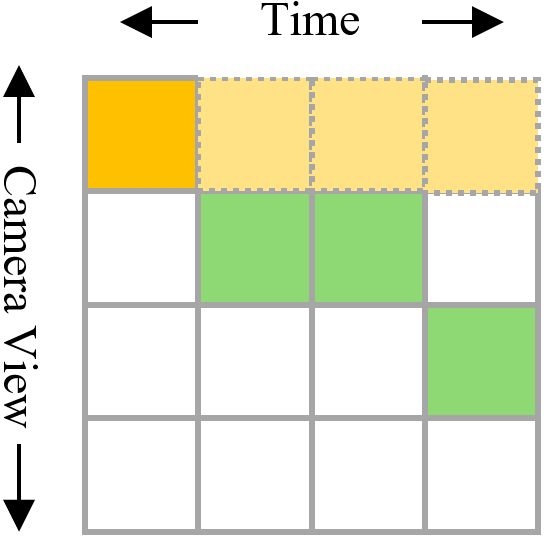} & 
    \includegraphics[width=2.7cm, valign=m]{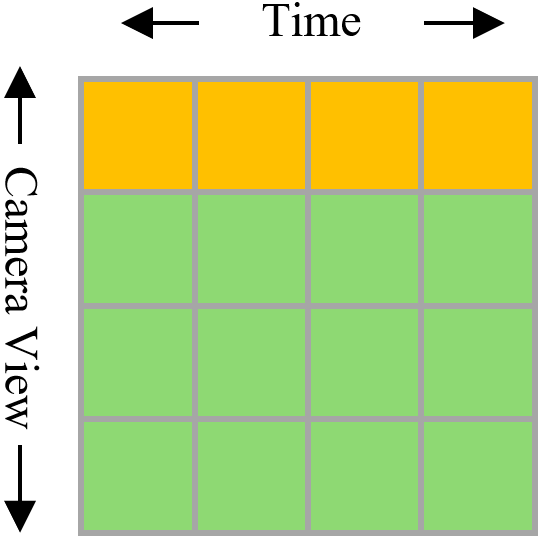} & 
    \includegraphics[width=2.7cm, valign=m]{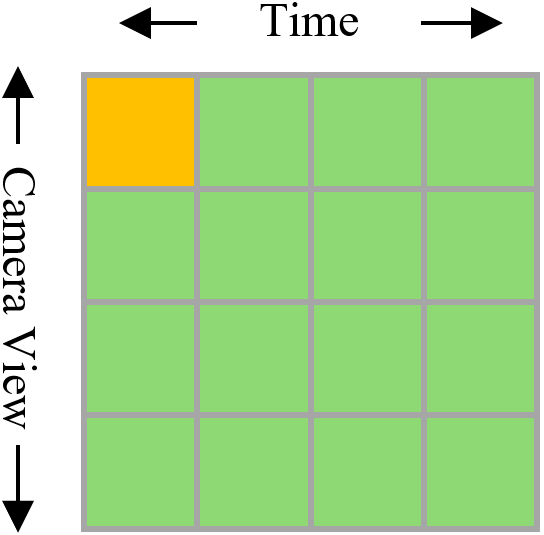} & 
    \includegraphics[width=2.7cm, valign=m]{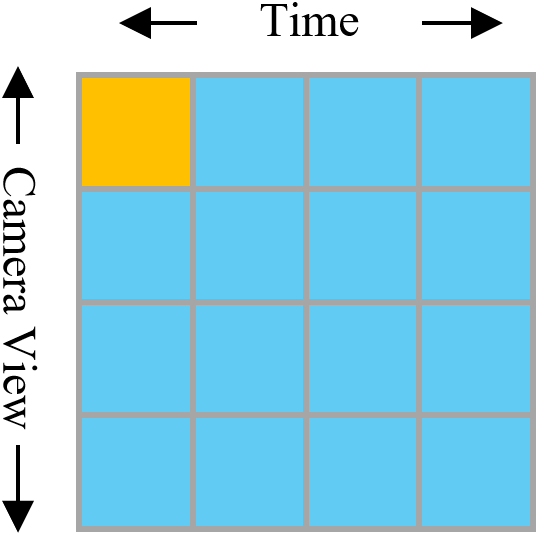} \\ \midrule

    Methods & Uni3C~\cite{cao2025uni3c}, DaS~\cite{gu2025diffusion}, MV-Custom~\cite{shin2026mvcustom}, NeoVerse~\cite{yang2026neoverse}, WorldStereo~\cite{zhang2026worldstereo}, WorldForge~\cite{song2025worldforge}, Gen3C~\cite{ren2025gen3c}, ReCamMaster~\cite{Bai_2025_ICCV}, Viewcrafter~\cite{yu2024viewcrafter}, TrajectoryCrafter~\cite{mark2025trajectorycrafter}, Free4D~\cite{liu2025free4d}, \textit{etc.} & SV4D 2.0~\cite{Yao_2025_ICCV}, MV-Performer~\cite{zhi2025mv} & SynCamMaster~\cite{ICLR2025_9232d474}, CAT4D~\cite{wu2025cat4d} & HarmoHOI (Ours) \\ \midrule
    
    Generative Capacity & Restricted (Warp-and-inpaint) & Limited (Reconstruction-focused) & High (Appearance-only synthesis) & High (Geometry-aware synthesis) \\ \midrule
    
    3D Modeling & Implicit (Mostly 2D warped video) & Dependent (Mined from source views) & Absent (Appearance-only) & Explicit (Joint 2D-3D modeling) \\ \midrule
    
    View Consistency & Trajectory-limited (Single-path) & Inconsistent (Lacks global alignment) & Implicit (Data-driven synchronization) & Strict (Globally aligned 3D motion) \\ \bottomrule
    \end{tabularx}
\end{table*}

\section{Additional Implementation Details.}
\label{sec:add_implement}

Our HarmoHOI is built upon the open-source pretrained text-to-video model WAN~2.1-1.3B-T2V~\cite{wan2025wan}. The model contains 30 DiT modules, each with 12 attention heads, where the hidden dimension of each head is 128 and the total hidden dimension is 1536. The maximum text token length is set to 512. To reduce computational cost, we adopt a video resolution of 49 $\times$ 256 $\times$ 384. The VAE in WAN~2.1 uses spatiotemporal compression ratios of $(8 \times 8 \times 4)$, and its visual tokenizer further performs an additional $2\times$ downsampling. To recover global metric scales, we introduce two learnable scale tokens. To extend the model to conditional video generation with reference images and camera poses, we concatenate the epipolar rendered reference image tokens and the tokenized camera pl\"ucker map features with the video tokens along the temporal dimension. The input of the intra-view Spatiotemporal Attention module therefore consists of image tokens, camera tokens, and video tokens, leading to a total token length of $(2 + (\lfloor 49/4\rfloor + 1)) \times (256/16) \times (384/16) = 5760$. The input of the inter-view Geometric Attention module consists of camera tokens, scale tokens, and video tokens, resulting in a total token length of $(1 + 2 + 6) \times (256/16) \times (384/16) = 3456$. Our Global Motion Aligning Diffusion, GloMAD, is built upon Point Transformer V3~\cite{wu2024point} with sparse convolution. During training, we first warm up the WAN~2.1 foundation model for 5K steps using 10\% of the HOIGen1M~\cite{liu2025hoigen} data under the text and image conditioned video generation task. We then conduct full training following our hybrid-data multi-stage curriculum strategy. The training is performed on 8 NVIDIA A100 80GB GPUs, and we adopt memory optimization strategies including DeepSpeed ZeRO-3~\cite{rajbhandari2020zero}, gradient checkpointing, and mixed precision training.

\textbf{Computational Cost.}
Tab.~\ref{tab:comp_cost} compares model size and peak training memory under a unified setting (batch size~$1$, gradient accumulation steps~$8$). HarmoHOI uses $2.68$B parameters and $42.3$\,GB VRAM, remaining competitive with multi-view baselines while being substantially lighter than SViMo~\cite{dang2025svimo} and DaS~\cite{gu2025diffusion}.

\begin{table}[!htbp]
\centering
\caption{Model size and training-memory comparison.}
\label{tab:comp_cost}
\resizebox{\linewidth}{!}{%
\begin{tabular}{lccccc}
\toprule
 & SViMo & DaS & SV4D~2.0 & SynCamMaster & HarmoHOI \\
\midrule
Params.\ (B) & 5.82 & 8.13 & 2.01 & 1.77 & 2.68 \\
Training VRAM (GB) & 61.5 & 66.4 & 67.2 & 37.4 & 42.3 \\
\bottomrule
\end{tabular}}
\end{table}

\textbf{Coverage of Depth-aware Rendering.}
For HOIGen1M~\cite{liu2025hoigen}, Depth Anything~3~\cite{lin2025depth} estimates full-image depth from monocular videos; we deliberately avoid restricting depth to segmented hand--object regions so as not to introduce additional segmentation errors. For TACO~\cite{liu2024taco}, depth is rendered from ground-truth 3D meshes and poses. During in-the-wild inference, the same full-image depth estimation is applied to the reference image for depth-aware epipolar rendering (Alg.~\ref{alg:motion_rep}).

\section{Additional Evaluation Details}
\label{sec:add_eval}

\textbf{Physical Plausibility and Perceptual Evaluation.}
We assess 3D-motion plausibility following SViMo~\cite{dang2025svimo}, using SAM~3 to segment hand and object regions and reporting penetration and non-contact rates in Tab.~\ref{tab:motion_quality}.
To further evaluate perceptual preference, we conduct a user study on $10$ motion cases with $37$ valid responses.
As shown in Tab.~\ref{tab:user_study_motion}, participants overwhelmingly prefer HarmoHOI over Geo4D~\cite{jiang2025geo4d}, GeometryCrafter~\cite{xu2025geometrycrafter}, and Depth Anything~3~\cite{lin2025depth}.
For RGB videos, qualitative comparisons are provided in the supplementary video (4:10--5:48).
A second study on $10$ six-view cases (likewise $37$ valid responses) yields a similarly strong preference for HarmoHOI over DaS~\cite{gu2025diffusion}, SV4D~2.0~\cite{Yao_2025_ICCV}, and SynCamMaster~\cite{ICLR2025_9232d474} (Tab.~\ref{tab:user_study_video}).

\begin{table}[!htbp]
\centering
\caption{User-study preference (\%) on 3D motion quality.}
\label{tab:user_study_motion}
\resizebox{\linewidth}{!}{%
\begin{tabular}{lcccc}
\toprule
 & Geo4D & GeometryCrafter & DA~3 & HarmoHOI \\
\midrule
Preference (\%) & 0.00 & 0.00 & 2.43 & \textbf{97.57} \\
\bottomrule
\end{tabular}}
\end{table}

\begin{table}[!htbp]
\centering
\caption{User-study preference (\%) on multi-view RGB video quality.}
\label{tab:user_study_video}
\resizebox{\linewidth}{!}{%
\begin{tabular}{lcccc}
\toprule
 & DaS & SV4D~2.0 & SynCamMaster & HarmoHOI \\
\midrule
Preference (\%) & 1.09 & 4.05 & 0.00 & \textbf{94.86} \\
\bottomrule
\end{tabular}}
\end{table}

\textbf{PI Metric.}
Following GeometryCrafter~\cite{xu2025geometrycrafter}, a predicted point $\hat{p}$ is counted as an inlier if
\begin{equation}
\frac{\|\hat{p}-p\|_2}{\min(\|p\|_2,\|\hat{p}\|_2)} < 0.25.
\end{equation}
The single- versus multi-view gap in Tab.~\ref{tab:motion_quality} arises from additional cross-view scale and alignment errors, together with inter-view visibility and point-coverage differences.

\section{Additional Demonstrations}

In this work, we focus on the task of image to video and motion generation, aiming to synthesize temporally coherent videos together with corresponding motion representations from a given reference image. To provide a more vivid, intuitive, and comprehensive presentation of our results, we include a \textbf{supplementary video} in the appendix. This video presents a diverse set of generated examples produced by our method, covering different interaction scenarios, object categories, motion patterns, and camera viewpoints. In addition, we provide systematic visual comparisons with several representative baseline approaches, allowing readers to better assess the generation quality from both appearance and motion perspectives. These comparisons show that our method produces more realistic visual details, more plausible interaction dynamics, and more consistent results across multiple views. Overall, the supplementary video serves as an important complement to the quantitative results and further demonstrates the effectiveness and generalization capability of our approach.

\end{document}